%% file: templateArxiv.tex
\documentclass{article}

\usepackage{PRIMEarxiv}

\usepackage[utf8]{inputenc} 
\usepackage[T1]{fontenc}    
\usepackage{hyperref}       
\usepackage{url}            
\usepackage{booktabs}       
\usepackage{amsfonts}       
\usepackage{nicefrac}       
\usepackage{microtype}      
\usepackage{lipsum}
\usepackage{fancyhdr}       
\usepackage{graphicx}       
\graphicspath{{media/}}     

\let\svthefootnote\thefootnote
\newcommand\freefootnote[1]{%
  \let\thefootnote\relax%
  \footnotetext{#1}%
  \let\thefootnote\svthefootnote%
}

\title{Self-Supervised Monocular Depth Estimation of
Untextured Indoor Rotated Scenes}

\author{
  Benjamin Keltjens \\
  Delft University of Technology \\
  \texttt{benjaminkeltjens@gmail.com} \\
   \And
  Tom van Dijk \\
  Delft University of Technology \\
  \texttt{J.C.vanDijk-1@tudelft.nl} \\
  
  \And
  Guido C.H.E. de Croon \\
  Delft University of Technology \\
  \texttt{g.c.h.e.decroon@tudelft.nl} \\
}

\def\etal{\emph{et al}.}

\begin{document}
\maketitle

\begin{abstract}
Self-supervised deep learning methods have leveraged stereo images for training monocular depth estimation. Although these methods show strong results on outdoor datasets such as KITTI, they do not match performance of supervised methods on indoor environments with camera rotation. Indoor, rotated scenes are common for less constrained applications and pose problems for two reasons: abundance of low texture regions and increased complexity of depth cues for images under rotation. In an effort to extend self-supervised learning to more generalised environments we propose two additions. First, we propose a novel Filled Disparity Loss term that corrects for ambiguity of image reconstruction error loss in textureless regions. Specifically, we interpolate disparity in untextured regions, using the estimated disparity from surrounding textured areas, and use L1 loss to correct the original estimation. Our experiments show that depth estimation is substantially improved on low-texture scenes, without any loss on textured scenes, when compared to Monodepth by Godard et al. Secondly, we show that training with an application's representative rotations, in both pitch and roll, is sufficient to significantly improve performance over the entire range of expected rotation. We demonstrate that depth estimation is successfully generalised as performance is not lost when evaluated on test sets with no camera rotation. Together these developments enable a broader use of self-supervised learning of monocular depth estimation for complex environments.
\end{abstract}

\freefootnote{The code for this work used can be found at \url{https://github.com/tudelft/filled-disparity-monodepth}}
\freefootnote{The data used can be found at \url{https://dataverse.nl/dataset.xhtml?persistentId=doi\%3A10.34894\%2FKIBWFC}}


\input{sections_new/introduction}

\input{sections_new/related_work}
\input{sections_new/proposed_method}
\input{sections_new/results}
\input{sections_new/conclusions}

\bibliographystyle{unsrt}  
\bibliography{references}  

\newpage
\appendix
\input{appendix/Data_Collection}

\input{appendix/Computational}
\newpage
\input{appendix/Additional_Rotations}

\end{document}

%% file: sections_new/introduction.tex
\section{Introduction}
\label{sec:intro}

The ability to estimate depth in a scene is an essential component of many 3D computer vision tasks. It is extremely useful for many applications, such as scene reconstruction, autonomous navigation of indoor environments, augmented reality and image augmentation. \par

Initially, depth estimation was done through classical techniques such as stereo matching \cite{SGM_2005} \cite{graph-cut} and Structure from Motion (SfM) \cite{sfm}, amongst others. More recently, methods using Convolutional Neural Networks (CNNs) have been developed for the purpose of estimating depth from a single image \cite{eigen} \cite{Garg2016} \cite{left-right} \cite{Godard2019} \cite{Zhou2017}, opening the possibility to more lightweight and flexible robots and devices having this capability. This was first achieved using ground-truth labels for depth to train in a supervised fashion \cite{eigen} \cite{Jung2017}. The accuracy of supervised methods is impressive on datasets like KITTI \cite{KITTI}, featuring outdoor car scenes, and NYUDepthV2 \cite{NYU}, showing indoor scenes. However, supervised learning requires complex and cumbersome data collection methods, such as heavy LIDAR systems.

To combat this, self-supervised learning of monocular depth has emerged to guide learning using stereo cameras \cite{Garg2016} \cite{left-right} or monocular sequences of images \cite{Zhou2017}, where training data is more easily collected. These newer methods are promising for the application of online-learning and show strong performance on similar outdoor domains as supervised learning. However, stereo pair methods are generally applied to domains quite fixed in content and motion. We would like such networks to work in more generic settings, for example in indoor environments or when the camera is under rotation. These conditions still cause problems that hamper their widespread applicability \cite{Dijk2019}.

For less physically constrained applications on dynamic platforms, such as hand-held phones or autonomous Micro Air Vehicles (MAVs), indoor scenes under rotation are common. Indoor domains often feature untextured scenes with structures such as walls and ceilings. Self-supervised learning with image-reconstruction is based on estimating the disparity shift between stereo pairs to synthesise the alternate view and compare it to ground truth view of the other camera. This method is ill-posed for textureless regions as small or large translations of pixels are indistinguishable in the generated, alternate, view.

Additionally, these platforms must handle a wide range of difficult rotations which are challenging. Van Dijk and De Croon show that Monodepth \cite{left-right} and other networks have difficulties estimating depth under pitch and roll rotation when trained on KITTI \cite{Dijk2019}. This is explained by the dependence on the vertical position of objects as a depth cue for a camera with a fixed pose.

To improve the application of self-supervised learning in these more complex domains, we present two additions to the current state-of-the-art. First, we propose a novel loss term, Filled Disparity Loss, to learn correct disparity for textureless regions. We make use of an assumption, inspired by stereo matching disparity refinement \cite{ELAS}, that disparity can be interpolated between edges, where the image-reconstruction loss is better posed. Our method improves depth estimation considerably when compared to previous image-reconstruction based self-supervised networks. Secondly, we show that building a dataset with representative rotations expected in the network's deployment significantly improves performance without requiring any changes to the network's structure. These two findings allow self-supervised networks to be deployed in the studied complex, but common, domains.

%% file: sections_new/related_work.tex
\section{Related Work}
\label{sec:related_work}
Several advances have been made to estimate depth with monocular vision. Initially, hand-crafted features combined with probabilistic methods were developed, such as Saxena \etal's method which used Markov Random Fields \cite{Saxena2005}. However, more recently deep learning methods have come to the forefront to tackle the problem \cite{eigen} \cite{Garg2016} \cite{left-right} \cite{Godard2019} \cite{Zhou2017} .

\vspace{-5mm}
\paragraph{Supervised Monocular Depth Estimation}
Eigen \etal \cite{eigen} were the first to use deep learning to leverage learned features for monocular depth estimation. Their network used a two-stage process with coarse and refined depth estimation. Since this work, additions such as GANs \cite{Kumar2018} and multi-task learning have improved the performance of supervised methods. Some networks leverage semantic and surface normal labels to aid in depth estimation \cite{Eigen2015}. Performance is strong in various environments with datasets such as KITTI \cite{KITTI} for outdoor scenes for driving cars and NYU-Depth V2 \cite{NYU} that has a variety of indoor scenes. Supervised methods are able to handle textureless surfaces in indoor environments \cite{Fu2018} as depth is directly learned. To an extent, supervised methods have also been able to handle scenes with camera rotation, such as NYUDepth-V2 that contains slightly pitched images. Recently, Zhao \etal \cite{ZhaoPose2020} use an encoding of the camera pose as an additional input to significantly improve performance on rotated datasets.

\vspace{-5mm}
\paragraph{Self-Supervised Monocular Depth Estimation}
In order to avoid the vast amount of required labelled data for supervised methods, various self-supervised methods have been developed using both stereo images and monocular video sequences. Garg \etal developed the first self-supervised method using image reconstruction with stereo pairs \cite{Garg2016}. Instead of estimating depth directly, their proposed method estimates disparity, which is then used to warp the input left image to synthesise the right image of the stereo pair. This warped version is compared to the true right image, and a loss based on the difference between the two (the image reconstruction loss) is used to guide learning. Disparity, $d$, is converted to depth, $D$, by $D = \frac{bf}{d}$ where $b$ is the baseline distance and $f$ is focal length in pixels. Godard \etal \cite{left-right} improved on this method with the addition of a fully differentiable warping method as well as estimating the disparity for both the left and right image and defining a loss for consistency between them. There have been various improvements to these methods with diverse network structures as well as novel loss terms \cite{Godard2019} \cite{CPU} \cite{refine_distill}. In addition to self-supervision with stereo-images, methods have also been developed to leverage monocular videos as training material. Zhou \etal \cite{Zhou2017} proposed a method that simultaneously estimates pose and depth, allowing images to be warped between neighbouring frames and used for an image reconstruction loss. \par

Contrary to supervised methods, not much has been done to evaluate self-supervised learning with stereo images when training in indoor environments under rotation; the KITTI dataset is used most often for training and evaluation \cite{Garg2016} \cite{left-right} \cite{refine_distill}. This leads to degradation of performance when testing on rotated images and datasets \cite{Dijk2019}. Closest to our work is that of Yu \etal \cite{Yu2020}. They use plane fitting of superpixel regions for self-supervised learning with monocular indoor image sequences. Their method assumes physical consistency within untextured regions, whereas ours assumes consistency of untextured regions with surrounding areas as well. Moreover, in contrast to \cite{Yu2020}, our method corrects all untextured areas and does not require an additional segmentation algorithm.


%% file: sections_new/proposed_method.tex
\section{Proposed Method}
\label{sec:proposed_method}

In order to estimate depth in indoor, rotated environments, our proposed method entails both a novel loss term using filled disparity maps as well as the use of representative datasets to learn expected rotations in environments. First, the losses and network are presented and then a description of the rotated datasets used for training is given.

\subsection{Filled Disparity Loss}

\begin{figure}[h]
    \centering
    \includegraphics[width=0.9\textwidth]{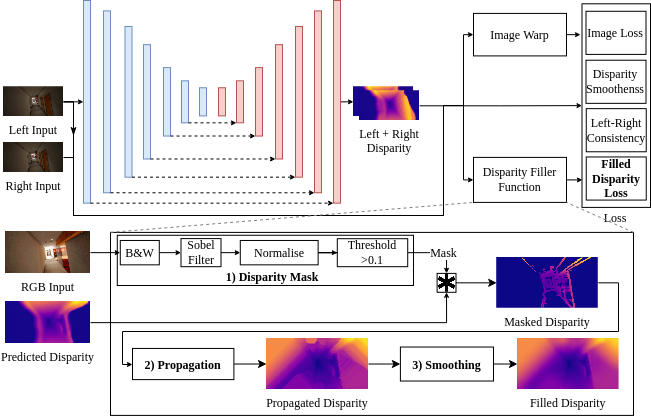}
    \caption{Structure of our network and the Disparity Filler Function}
    \label{fig:untextured_diagram}
\end{figure}

As described in \autoref{sec:related_work}, self-supervised learning of monocular depth using stereo images infers disparity maps from a single RGB image.  However, the image reconstruction loss in textureless regions of the scene is ill-posed due to small and large disparity maps producing the same result, and therefore loss, in those regions. The effect can be seen for the state-of-the-art Monodepth method \cite{left-right} in the second column of \autoref{fig:untextured_examples}. We propose an additional loss term, called Filled Disparity Loss, which uses filled disparity maps in regions of low texture as an additional loss for indoor scenes. This is inspired by the interpolation of sparse of disparity estimates in ELAS \cite{ELAS}. This new loss is used together with the losses for stereo self-supervised learning used by Godard \etal in Monodepth \cite{left-right} as can be seen in the network overview in \autoref{fig:untextured_diagram}. 

The Disparity Filler Function generates an interpolated disparity map as seen in the bottom of \autoref{fig:untextured_diagram}. This function is split into three main parts. 

\vspace{-3mm}
\paragraph{Disparity Mask} First, a sparse disparity mask is generated by identifying significant edges where the reconstruction loss is well posed. This is done by means of a Sobel filter of size 7 on the grayscale input image. A large filter is used to avoid detecting smaller changes due to noise, that may appear in regions of lower texture. Feature detection methods such as ORB \cite{ORB} and SURF \cite{SURF} were considered, however, they produce far too few, scattered points for the following steps. Additionally, Sobel filters perform well in the Tensorflow framework \cite{Tensorflow}. The gradient image is then normalised to the interval $[0,1]$ and then an "active" mask of textured pixels is defined as values greater than 0.1. 

\begin{figure}[h]
    \centering
    \includegraphics[width=0.9\textwidth]{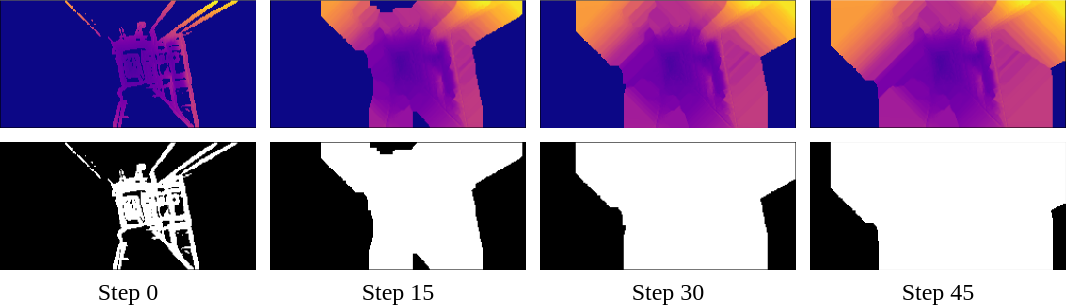}
    \caption{Progression of the Propagation step of the Disparity Filler Function to fill the sparse disparity map (top) and active pixel mask (bottom)}
    \label{fig:propagation}
\end{figure}
\paragraph{Propagation} Secondly, the disparity is propagated outwards to other edges. A while loop is used where pixels are filled by averaging the immediately neighbouring active pixels as can be seen in \autoref{fig:propagation}. This loop continues until the entire image has been filled with active pixels. This method allows for interpolation between edges as well as interpolation to the border of the image where there might be no active pixels.

\vspace{-5mm}
\paragraph{Smoothing} In the final step a 5x5 smoothing kernel, using inverse distance and 1 at the centre (normalised), is used to smoothen the new disparities in textureless regions. Textureless pixels are found using the inverse of the initial mask of active pixels.  

Overall, the Disparity Filler Function is able to generate more accurate disparity estimates in the textureless regions. However, from \autoref{fig:propagation} it is apparent that there are some undesired effects such as filled disparities being propagated perpendicular to detected edges. This can be improved by methods such as used by Geiger \etal \cite{ELAS} with interpolation over Delauny Triangulation between points. However, this process is not differentiable and performs slower on the CPU with large quantities of active pixels. The new map is compared to the originally estimated disparity map using an L1 loss.

\vspace{-3mm}
\subsection{Losses}
The loss function for this network is based on the work done by Godard \etal \cite{left-right} and consists of 4 terms, resolved at multiple scales $s$, as seen in \autoref{eq:general_losses}. The total loss is the sum of losses at 4 different scales $L = \sum_{s=1}^{4} L_{s}$.

\vspace{-5mm}
\begin{equation}
    L_{s} =  \alpha_{ap} \left( L_{ir}^{l} + L_{ir}^{r} \right) + \alpha_{ds} \left(L_{ds}^{l} + L_{ds}^{r} \right) + \alpha_{lr} \left(L_{lr}^{l} + L_{lr}^{r} \right) + \alpha_{fd} \left(L_{fd}^{l} + L_{fd}^{r}\right)
    \label{eq:general_losses}
\end{equation}
\vspace{-3mm}

$L_{ir}$ evaluates the loss of image reconstruction of input images, $L_{ds}$ evaluates the smoothness of the disparity map, $L_{lr}$ evaluates the consistency of the left and right disparity maps and $L_{fd}$ evaluates the similarity of the estimated disparity map and the filled disparity map. Disparities are predicted for both the left and right images; the superscript l and r indicate which image the loss is calculated for. The losses are shown in detail for the left images case.

\vspace{-5mm}
\paragraph{Image Reconstruction Loss} To learn the correct disparity map, both the left and right images are warped to match the opposite direction and are compared to the input images as described in \autoref{sec:related_work}. Similar to \cite{left-right} the imaging warping is done using bilinear sampling \cite{bilinear}. The reconstruction loss is measured by both L1 and SSIM, comparing input image $I_{ij}^{l}$ with the reconstruction $\tilde{I}_{ij}^{l}$, where $N$ is number of pixels and $\alpha = 0.85$ is the weighting term:

\begin{equation}
    L_{ir}^{l} = \frac{1}{N}\sum_{ij}^{}\alpha\frac{1-SSIM(I_{ij}^{l},\tilde{I}_{ij}^{l})}{2} + (1-\alpha)\left| I_{ij}^{l} - \tilde{I}_{ij}^{l} \right|
    \label{eq:SSIM}
\end{equation}

\vspace{-5mm}
\paragraph{Disparity Smoothness} This loss term encourages local smoothness by minimising gradients $\partial d$ in the disparity map. The loss takes account of large gradients at edges in the input image by reducing the loss with the image gradients $\partial I$.

\begin{equation}
    L_{ds}^{l} = \frac{1}{N}\sum_{ij}^{}\left| \partial_{x} d_{ij}^{l} \right|e^{-\left| \partial_{x}I_{ij}^{l} \right|}  + \left| \partial_{y} d_{ij}^{l} \right|e^{-\left| \partial_{y}I_{ij}^{l} \right|} 
    \label{eq:disparitysmoothness}
\end{equation}

\vspace{-5mm}
\paragraph{Left-Right Consistency} As the network outputs disparities for both the left and right images, this loss term ensures consistency between them. It uses the L1 loss between the left disparity map $d_{ij}^{l}$ and the right disparity map projected onto the left view $d_{ij+d_{ij}^{l}}$ \cite{left-right}.

\begin{equation}
    L_{lr}^{l} = \frac{1}{N}\sum_{ij}^{} \left| d_{ij}^{l} - d_{ij+d_{ij}^{l}} \right|
    \label{eq:LR}
\end{equation}

\vspace{-5mm}
\paragraph{Filled Disparity Loss} Our proposed loss is formulated as an L1 loss between the estimated disparity map $d_{ij}^{l}$ and the filled disparity map $\tilde{d}_{ij}^{l}$. The loss only applies to untextured regions as the Disparity Filling Function does not change disparity in textured regions of an image.

\begin{equation}
    L_{fd} = \frac{1}{N}\sum_{ij}^{} \left|d_{ij}^{l} - \tilde{d}_{ij}^{l} \right|
    \label{eq:propagated}
\end{equation}

\vspace{-6mm}
\subsection{Datasets}

This work makes use of the AirSim Building\_99 drone simulation environment developed by Microsoft \cite{airsim} to generate training data. A more detailed description of the collection method can be found in \autoref{appenidx:data_collection}. Various indoor scenes are captured with left and right images as well as ground truth disparity. The environment contains a range of scenes ranging from long narrow hallways to open spaces as seen in first column of \autoref{fig:untextured_examples}. Flying on a trajectory through the entire building, 18000 images (128x256) are collected for training and 2000 for testing. Depth is saturated on the boundaries 0-80 [m] as done in \cite{left-right}.

Four datasets are generated with pitch and roll motions enabled or disabled, as listed in \autoref{tab:rotation_datasets}. The angles are normally distributed in pitch and roll with a standard deviation of 10 degrees about 0 degrees.

\begin{table}[h]
\centering
\begin{tabular}{|c|c|}
\hline
Dataset ID & Dataset Description                                                                       \\ \hline
N          & Nominal with no rotations                                                                 \\ \hline
R          & Rotation in roll with standard deviation of 10 degrees                                  \\ \hline
P          & Rotation in pitch with standard deviation of 10 degrees                                 \\ \hline
PR         & Rotation in pitch and roll with standard deviation of 10 degrees \\ \hline
\end{tabular}
\caption{Dataset Descriptions}
\label{tab:rotation_datasets}
\end{table}

\vspace{-6mm}
\subsection{Implementation Details}
The hyperparameters of the network are the same as in \cite{left-right}. The network is trained for 50 epochs with a batch size of 8 using an Adam Optimiser where $\beta_{1}$, $\beta_{2}$, and $\epsilon$ are 0.9, 0.999 and $10^{-8}$ respectively. The learning rate $\lambda$ is $10^{-4}$ for 30 epochs and is then halved every 10 epochs. Colour augmentations and flipping of the input image pairs is also performed randomly for data augmentation, similar to \cite{left-right}. A detailed description of the computational performance on different platforms is given in \autoref{appenidx:computation}. 

%% file: sections_new/results.tex
\section{Results}
\label{sec:results}

To show our improvements we begin by analysing the effect of the Filled Disparity Loss on the performance on the overall dataset. We also evaluate the effect of image texturedness on performance. This is followed by an analysis of the effect of training with datasets that contain representative rotations on overall performance, as well as over the range of rotations.

\paragraph{Metrics} We use the same metrics as described in \cite{left-right}. Given ground truth depth $D^{*}$ and estimated depth $D$, the following metrics are defined. (1) Absolute Relative Error: $\frac{1}{\left| N \right|}\sum_{i\in N}^{}\frac{\left| D_{i} - D_{i}^{\ast} \right|}{D_{i}^{\ast}}$, (2) Squared Relative Error: $\frac{1}{\left| N \right|}\sum_{i\in N}^{}\frac{\left\|  D_{i} - D_{i}^{\ast} \right\|^2}{D_{i}^{\ast}}$, (3) RMSE: $\sqrt{\frac{1}{\left| N \right|}\sum_{i\in N}^{}\left\| D_{i} - D_{i}^{\ast} \right\|^{2}}$, (4) RMSE log: $\sqrt{\frac{1}{\left| N \right|}\sum_{i\in N}^{}\left\| log\left(D_{i}\right) - log\left(D_{i}^{\ast}\right) \right\|^{2}}$, (5) $\delta_{t}$: \% of $D_{i} \ s.t. \ max(\frac{D_{i}}{D_{i}^{\ast}}, \frac{D_{i}^{\ast}}{D_{i}}) < t$, where $t \in \left\{ 1.25, 1.25^{2}, 1.25^{3} \right\}$.

\paragraph{Performance Metrics on Untextured Scenes}

The effect of the new loss term can be seen by altering its weight $\alpha_{fd}$. The results displayed in \autoref{tab:untextured_performance} are from training and testing on the pitch and roll (PR) dataset. We can see that increasing $\alpha_{fd}$ leads to an improvement in almost all error metrics, except $\delta < 1.25$ at $\alpha_{fd} = 0.6$.  Without the new loss function, the network performs considerably worse when trained and tested on this domain. The decrease in the errors stagnates at larger values of the loss weight as it reaches the optimum. 

\begin{table}[h]
\centering
\scriptsize
\begin{tabular}{|c|cccc|ccc|}
\hline
\multicolumn{1}{|c|}{{$\alpha_{fd}$}} & \multicolumn{4}{c|}{Lower is better}                                                                                       & \multicolumn{3}{c|}{Higher is better}                          \\ \cline{2-8} 
                    & \multicolumn{1}{c|}{Abs. Rel.} & \multicolumn{1}{c|}{Sq. Rel.} & \multicolumn{1}{c|}{RMSE} & \multicolumn{1}{c|}{RMSE log} & \multicolumn{1}{c|}{$\delta < 1.25$} & \multicolumn{1}{c|}{$\delta < 1.25^{2}$} &   $\delta < 1.25^{3}$    \\ \hline
0.0 (Monodepth)                                             & 1.7676                         & 89.3917                       & 13.692                    & 0.707                         & 0.667                 & 0.782                 & 0.840          \\
0.1                                          & 1.1126                         & 49.5702                       & 10.575                    & 0.572                         & 0.685                 & 0.803                 & 0.865          \\
0.2                                          & 0.3696                         & 2.8818                        & 6.301                     & 0.417                         & \textbf{0.687}        & 0.816                 & 0.882          \\
0.3                                          & 0.3262                         & 2.3222                        & 6.175                     & 0.396                         & 0.683                 & \textbf{0.819}        & 0.892          \\
0.4                                          & \textbf{0.3172}                & 2.1787                        & 6.201                     & \textbf{0.392}                & 0.676                 & \textbf{0.819}        & \textbf{0.895} \\
0.5                                          & 0.3200                         & 2.1690                        & \textbf{6.109}            & \textbf{0.392}                & 0.674                 & \textbf{0.819}        & \textbf{0.895} \\
0.6                                          & 0.3181                         & \textbf{2.1549}               & 6.246                     & 0.394                         & 0.664                 & 0.815                 & 0.894          \\ \hline
\end{tabular}
\caption{Performance of networks with changing $\alpha_{fd}$ on the PR dataset.}
\label{tab:untextured_performance}
\end{table}

\begin{figure}[h]
    \centering
    \includegraphics[width = 0.9\textwidth]{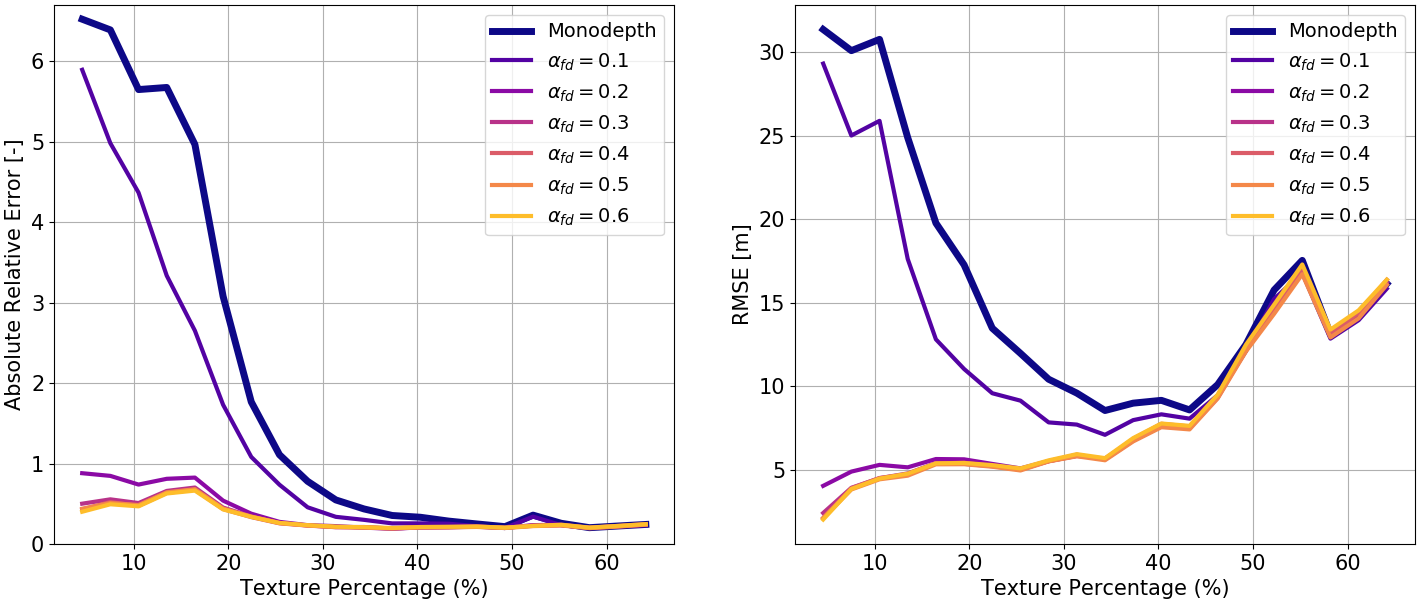}
    \caption{Absolute Relative error (left) and RMSE (right) of the PR test set as a function of the percentage of textured pixels in an image for varying $\alpha_{fd}$. Lower is better.}
    \label{fig:untextured_comparison}
\end{figure}

\paragraph{Influence of Texturedness on Performance}
We are interested in seeing the effect of texturedness on the network's performance. Texturedness is measured by the percentage of pixels with a Sobel gradient above our threshold of 0.1. In \autoref{fig:untextured_comparison} it is apparent that for lower levels of texture the error is much higher for networks trained with a lower loss weight $\alpha_{fd}$. As the texturedness of an image increases, the errors converge to the same value. This confirms that the new loss function significantly improves estimation in textureless regions and preserves performance in textured regions.

\begin{figure}[h]
    \centering
    \includegraphics[width=0.8\textwidth]{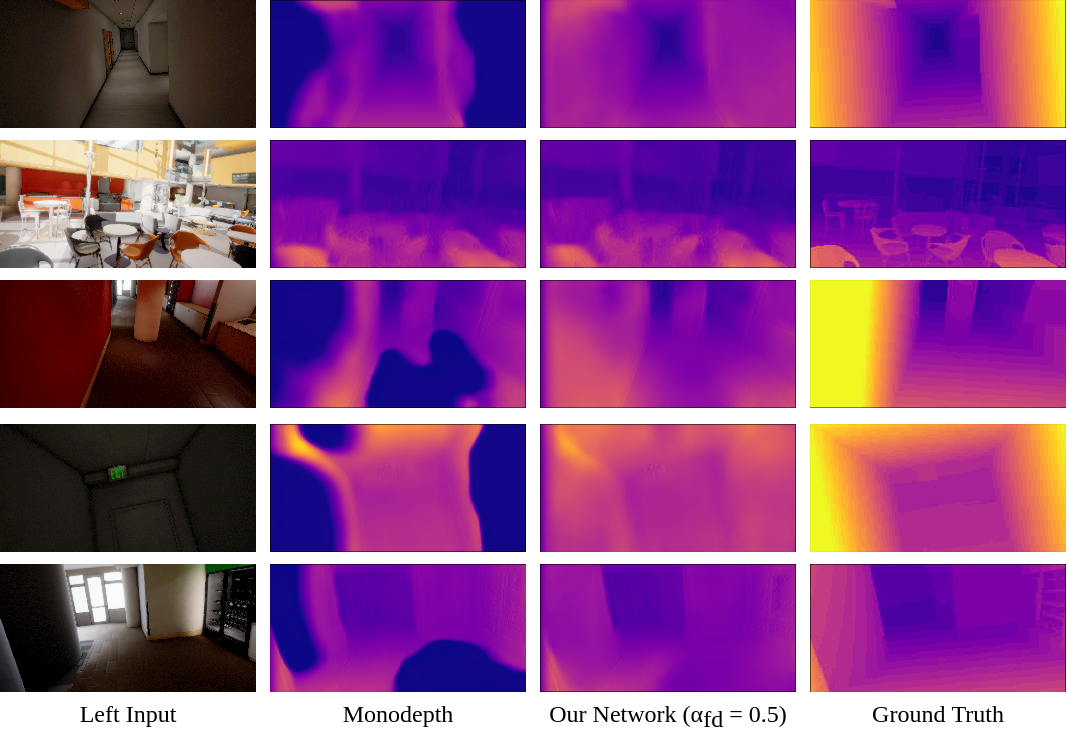}
    \caption{Disparity maps for different scenes on the PR test set}
    \label{fig:untextured_examples}
\end{figure}

\paragraph{Qualitative Performance on Untextured Scenes}
The qualitative results of the network with the Filled Disparity Loss can be seen in \autoref{fig:untextured_examples}. The errors from Monodepth are quite severe as disparity estimation in close-by, untextured, regions are extremely low (faraway regions). It is apparent that the additional loss term has helped to guide the network in textureless areas to correct for the ill-posed image reconstruction. This is especially apparent in dark environments where texture is lost (row 4), which can be common indoors. Additionally, in textured regions where Monodepth performs well, it is apparent that the performance is not degraded. Even though there is large improvement, our network still does not completely capture the scale of disparities in near, untextured, regions (rows 1, 3, 4). However, when the disparity is inverted to find depth, the differences at close range are small. \par
 
\paragraph{Performance Metrics on Rotated Datasets}
Our results in \autoref{tab:rotation_performance} show the effect of training on the different datasets when tested on the nominal (N) and pitch and roll (PR) datasets. The networks are all trained with $\alpha_{fd} = 0.5$. The first result of interest is that training the networks on datasets that are rotated does not seem to significantly affect the performance on the nominal dataset. \par


\begin{table}[h]
\centering
\scriptsize
\begin{tabular}{|c|c|cccc|ccc|}
\hline
\multicolumn{1}{|c|}{{\begin{tabular}[c]{@{}c@{}}Training\\ Dataset\end{tabular}}} & \multicolumn{1}{c|}{{\begin{tabular}[c]{@{}c@{}}Testing\\ Dataset\end{tabular}}} & \multicolumn{4}{c|}{Lower is better}                                                                                       & \multicolumn{3}{c|}{Higher is better}                          \\ \cline{3-9} 
\multicolumn{1}{|c|}{}                                                                            & \multicolumn{1}{c|}{}                                                                           & \multicolumn{1}{c|}{Abs. Rel.} & \multicolumn{1}{c|}{Sq. Rel.} & \multicolumn{1}{c|}{RMSE} & \multicolumn{1}{c|}{RMSE log} & \multicolumn{1}{c|}{$\delta < 1.25$} & \multicolumn{1}{c|}{$\delta < 1.25^{2}$} &   $\delta < 1.25^{3}$               \\ \hline
N                                                                                                 & N                                                                                               & 0.3156                         & 2.3348                        & 6.366                     & 0.394                         & 0.673                 & 0.815                 & 0.893          \\
R                                                                                                 & N                                                                                               & 0.3089                         & 2.1977                        & 6.424                     & 0.393                         & 0.670                 & 0.815                 & 0.894          \\
P                                                                                                 & N                                                                                               & 0.3134                         & 2.5294                        & 6.370                     & 0.391                         & 0.675                 & 0.819                 & \textbf{0.897} \\
PR                                                                                                & N                                                                                               & \textbf{0.3052}                & \textbf{2.1101}               & \textbf{6.287}            & \textbf{0.387}                & \textbf{0.679}        & \textbf{0.821}        & \textbf{0.897} \\ \hline \vspace{-2mm}
                                                                                                  &                                                                                                 &                                &                               &                           &                               &                       &                       &                \\ \hline
N                                                                                                 & PR                                                                                              & 0.7245                         & 7.4258                        & 10.232                    & 0.673                         & 0.317                 & 0.542                 & 0.701          \\
R                                                                                                 & PR                                                                                              & 0.4737                         & 3.9536                        & 7.992                     & 0.507                         & 0.483                 & 0.701                 & 0.821          \\
P                                                                                                 & PR                                                                                              & 0.3461                         & 2.3993                        & 6.697                     & 0.421                         & 0.611                 & 0.789                 & 0.880          \\
PR                                                                                                & PR                                                                                              & \textbf{0.3200}                & \textbf{2.1690}               & \textbf{6.109}            & \textbf{0.392}                & \textbf{0.674}        & \textbf{0.819}        & \textbf{0.895} \\ \hline
\end{tabular}
\caption{Performance of networks on the test sets of nominal (N) and Pitch \& Roll (PR)}
\label{tab:rotation_performance}
\end{table}

When tested on the pitch and roll test set, the discrepancy between the networks is noticeable, as seen in \autoref{tab:rotation_performance}. The network trained on the nominal dataset performs extremely poorly on the PR dataset when compared to the others and its performance on the nominal dataset. As expected, with more axes of rotation in the training dataset the network performs better. Training on the PR dataset has similar results when testing on the nominal and PR test sets. The network trained solely on pitch (P) performs significantly better than that trained only on roll (R). This is probably because training on pitch allows the learning of correct vertical height depth cues that arise from pitch rotation \cite{Dijk2019}.

\begin{figure}[h]
    \centering
    \includegraphics[width = 0.96\textwidth]{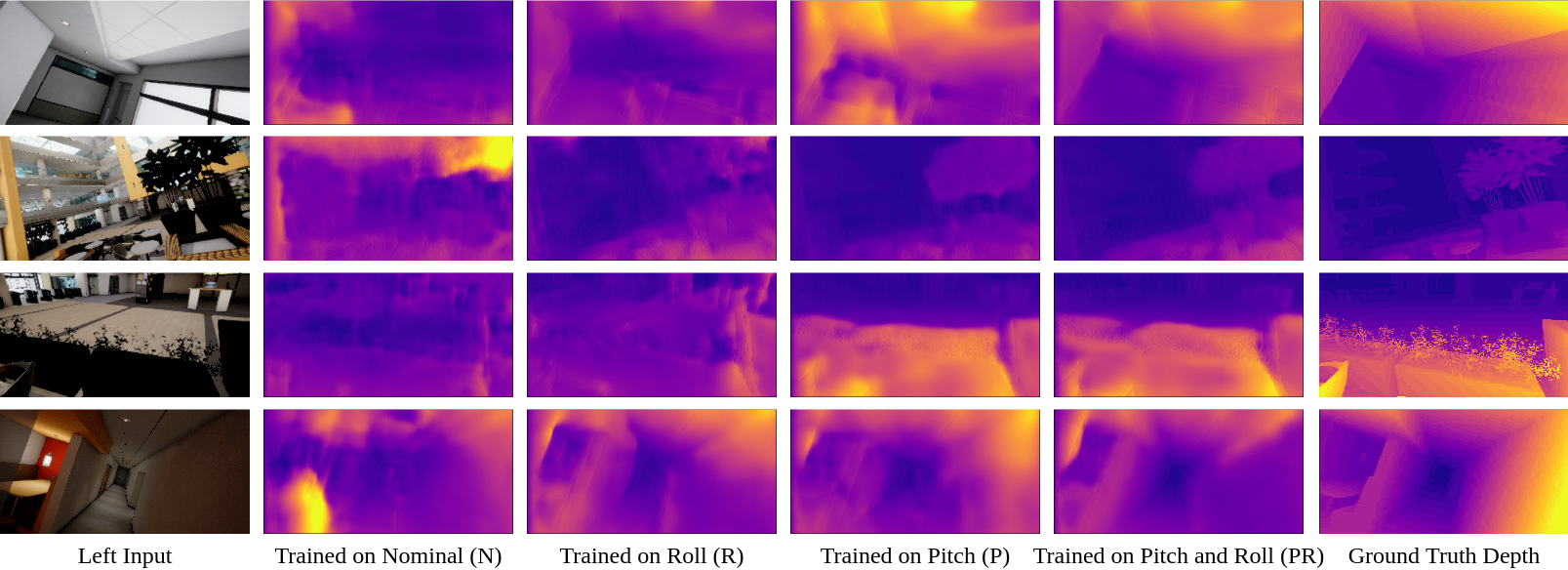}
    \caption{Comparison of disparity maps on PR test set. Networks trained with $\alpha_{fd} = 0.5$.}
    \label{fig:rotated_examples}
\end{figure}

\paragraph{Qualitative Performance on Rotated Datasets}
Looking at a comparison of disparity maps in \autoref{fig:rotated_examples} we can see that the performance of the network trained on the nominal dataset is severely degraded and struggles to retain any resemblance of structure when testing on rotated datasets. With more representative rotations in the training set the preservation of structure in the depth maps greatly improves. \par

\begin{figure}[h]
    \centering
    \includegraphics[width=0.79\textwidth]{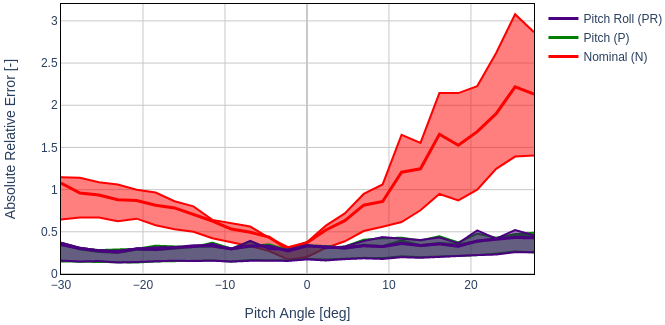}
    \caption{Comparison of networks trained on nominal (N), pitch (P) and pitch and roll (PR) datasets on the pitch (P) test set as a function of pitch angle. The 25th and 75th percentiles of the data are shown by the shaded areas. Pitch (P) and pitch and roll (PR) are almost identical.}
    \label{fig:rotated_pitch}
\end{figure}

\paragraph{Performance over Range of Rotation}
From \autoref{fig:rotated_pitch} it is evident that training on the pitch and roll (PR) dataset does not reduce performance on the pitch dataset (P) when compared to the network trained only on pitch (P). As expected, the error of the network trained on the nominal dataset (N) is much larger when the absolute value of the pitch angle increases. Also promising is that for the two networks trained with rotations, their performance remains quite constant over the range of pitch. The spread of error on the rotated dataset results are also much smaller than the network on nominal (N), indicating more consistent performance. Supplementary analysis for the roll (R) and the pitch and roll (PR) test sets are given in \autoref{appenidx:rotation}. \par

%% file: sections_new/conclusions.tex
\section{Conclusion}
\label{sec:conclusions} 
In this work we have presented a new Filled Disparity Loss term to improve depth estimation in textureless regions of images. Our method successfully estimates depth in untextured regions of indoor environments whilst preserving performance on textured regions.  Additionally, we have demonstrated the ability of self-supervised networks for monocular depth estimation to generalise over rotations of scenes given a representative dataset. Overall, this work allows for more mobile applications of self-supervised monocular depth estimation in complex, indoor, environments.  

%% file: appendix/Data_Collection.tex
\section{Data Collection}
\label{appenidx:data_collection}

\begin{figure}[h]
    \centering
    \includegraphics[width=0.9\textwidth]{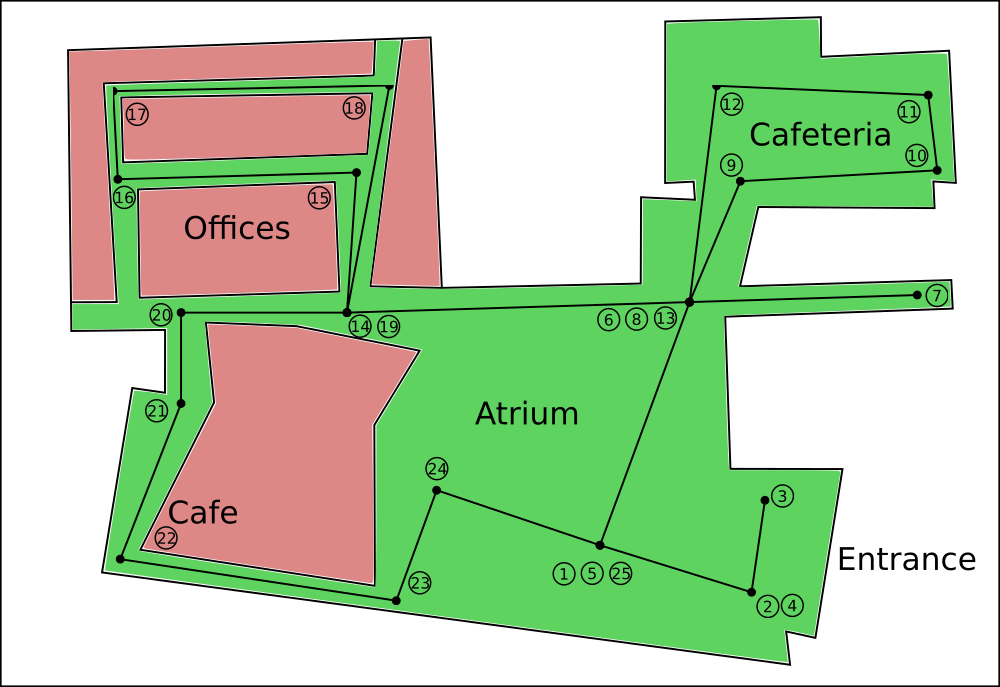}
    \caption{Floor plan of Building\_99 Simulation environment with labelled route}
    \label{fig:building99}
\end{figure}

\begin{figure}[h]
    \centering
    \includegraphics[width=0.9\textwidth]{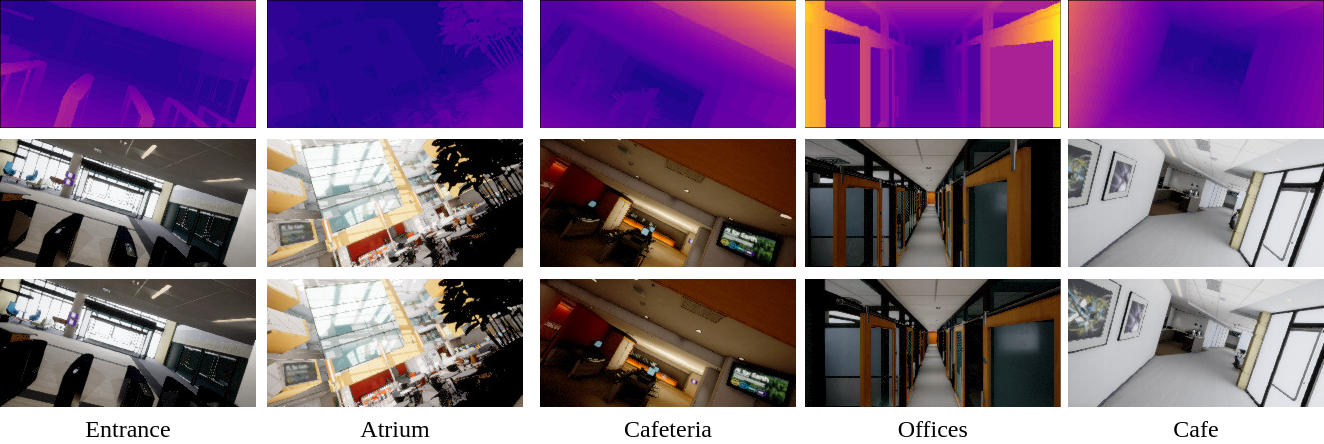}
    \caption{Example scenes of locations in \textit{Building\_99} under Pitch and Roll Rotation. Shown are the left image (bottom), right image (middle) and the ground truth disparity (top)}
    \label{fig:example_scenes}
\end{figure}

Data collection for this work was done in the Airsim \textit{Building\_99} simulation environment using binaries\footnote{The specific binary version can be found at \hyperlink{https://github.com/microsoft/AirSim/releases/tag/v1.3.1-linux}{https://github.com/microsoft/AirSim/releases/tag/v1.3.1-linux}} from Microsoft \cite{airsim}. The drone platform with stereovision and disparity ground truth for the left view was used to collect data in the environment. Movement of the drone was done by setting the exact pose of the drone as it allowed for control of the distribution of rotations. The translation of the drone followed a set path that allowed it to view the different parts of the building environment. The path taken over the building floor plan can be seen in \autoref{fig:building99}. Examples of the different scenes in \textit{Building\_99} can be seen in \autoref{fig:example_scenes}

%% file: appendix/Computational.tex
\section{Computational Performance}
\label{appenidx:computation}

The addition of the Filled Disparity Loss increases the computational load required for training when compared to Godard \etal \cite{left-right}. On a single GTX 1080 Ti Monodepth \cite{left-right} trains at 48.3 [examples/s] whilst our network achieves 29.5 [examples/s]. As the Filled Disparity function is only imposed in training, the speed in inference, 82.6 [examples/s], is the same for both. \par

In the interest of improving computational performance, a simple trimming process is performed where the number of layers in both the encoder and decoder is reduced to a percentage of the original size. The results of the training speed and accuracy on the Pitch and Roll (PR) dataset on three different platforms can be seen in \autoref{tab:computation_performance}. The three Nvidia platforms considered are the GTX 1080 Ti, Jetson TX2 and Jetson Nano. The smaller Jetson platforms are interesting in their ability to be placed on light mobile robots, such as drones, to enable online learning. \par

\begin{table}[h]
\centering
\scriptsize
\begin{tabular}{|c|c|c|c|c|c|c|c|c|}
\hline
{\begin{tabular}[c]{@{}c@{}} Network\\ Size {[}\%{]}\end{tabular}} & \multicolumn{3}{c|}{Training Speed {[}examples/s{]}} & \multicolumn{5}{c|}{Error Metrics}                                                 \\ \cline{2-9} 
                                                                                            & GTX 1080 Ti      & Jetson TX2      & Jetson Nano     & Abs. Rel.      & RMSE           & $\delta < 1.25$      & $\delta < 1.25^{2}$      & $\delta < 1.25^{3}$      \\ \hline
20                                                                                          & 58.80            & 6.67            & 1.81            & 0.348          & 6.689          & 0.632          & 0.795          & 0.879          \\ \hline
40                                                                                          & 55.13            & 5.88            & 1.68            & 0.331          & 6.382          & 0.656          & 0.808          & 0.887          \\ \hline
60                                                                                          & 43.98            & 5.08            & N/A             & 0.320          & 6.195          & 0.667          & 0.815          & 0.893          \\ \hline
80                                                                                          & 35.60            & 4.51            & N/A             & \textbf{0.313} & 6.119          & \textbf{0.677} & \textbf{0.821} & \textbf{0.896} \\ \hline
100                                                                                         & 29.50            & 3.86            & N/A             & 0.320          & \textbf{6.109} & 0.674          & 0.819          & 0.895          \\ \hline
\end{tabular}
\caption{Training speed and error metrics for differing network size on three Nvidia platforms. N/A is used to indicate that it was not possible to train the network for that size. Some error metrics are removed in the interest of space.}
\label{tab:computation_performance}
\end{table}

As expected, the training speed improves for a decrease in the size of the network on all platforms. On the Jetson Nano, only smaller sized networks can be trained due to memory constraints. Reducing the batch size did not enable training of the network. An interesting result in \autoref{tab:computation_performance} is that the performance in accuracy is not degraded by too large an amount whilst retaining only 20\% of the layers. \par

\begin{figure}[h]
    \centering
    \includegraphics[width=0.98\textwidth]{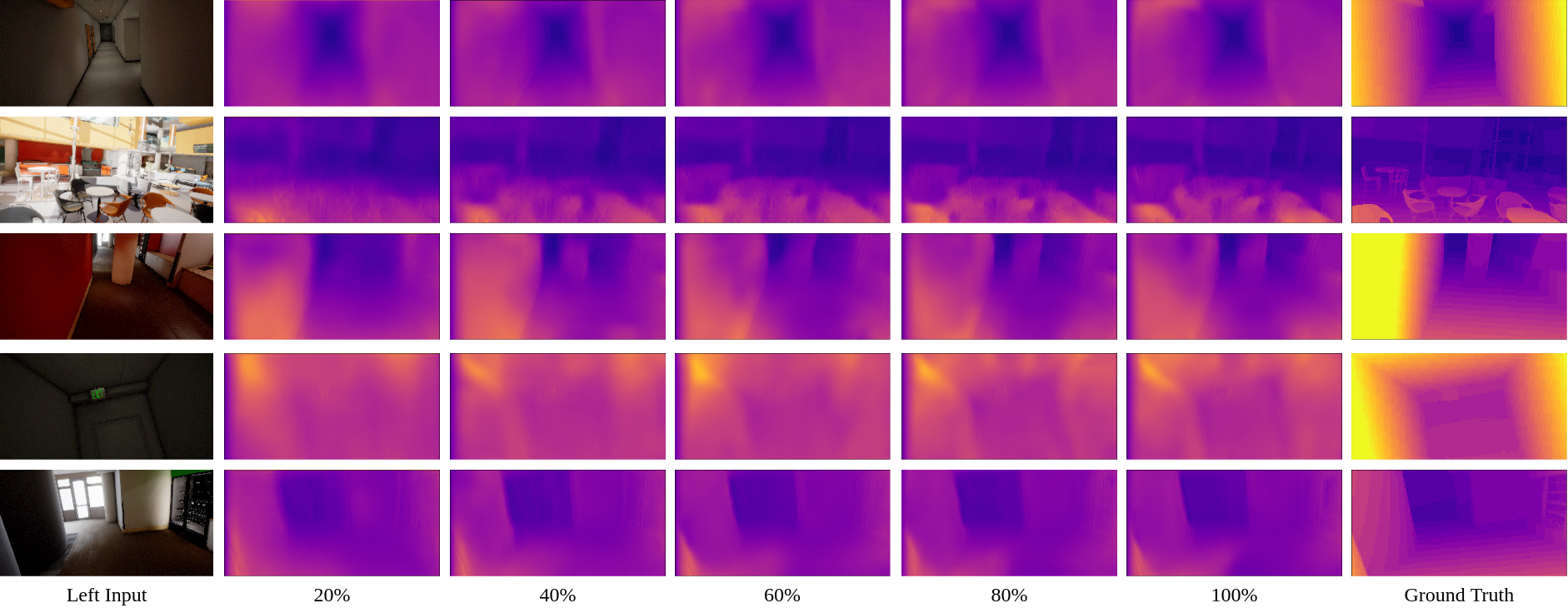}
    \caption{Comparison of disparity map outputs for networks with differing sizes. Sizes are given as percentage of nominal network size.}
    \label{fig:computation_comparison}
\end{figure}

\autoref{fig:computation_comparison} shows a qualitative comparison of the different network sizes. The scale of the disparity estimation is quite constant and the performance of the Filled Disparity Loss is retained in untextured areas. Though it does seem that detail is lost in more textured areas. For example, in the second row, the chairs which are delineated quite well in the original 100\% sized network become a single blur in the network 20\% the original's size. This seems promising for producing coarse estimations on low-power and light platforms.

%% file: appendix/Additional_Rotations.tex
\section{Supplementary Rotational Analysis}
\label{appenidx:rotation}

In addition to the rotational analysis given in \autoref{sec:results} a few additional results are presented here. First, an analysis of the effect of the roll angle on performance on the roll (R) test set is given followed by a discussion of the effect on performance over both the pitch and roll angle on the pitch and roll (PR) test set.\par

\begin{figure}[h]
    \centering
    \includegraphics[width=0.9\textwidth]{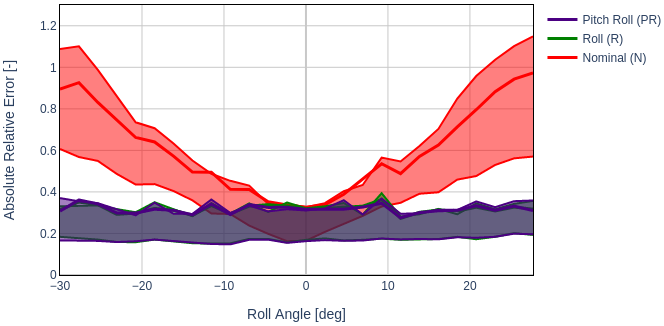}
    \caption{Comparison of networks trained on nominal (N), roll (R) and pitch and roll (PR) datasets on the roll (R) test set as a function of roll angle. The 25th and 75th percentiles of the data are shown by the shaded areas. Roll (R) and pitch and roll (PR) are almost identical.}
    \label{fig:roll_comparison}
\end{figure}

In \autoref{fig:roll_comparison} a comparison of the performance on the roll (R) test set is shown for 3 different networks (trained on nominal (N), roll (R) and pitch and roll (PR)). As expected the results are quite similar to that of \autoref{fig:rotated_pitch}. The performance of the network trained on the nominal (N) dataset is similar to the others for low roll angles but steadily gets worse for larger roll angles. Contrary to pitch, the decrease in performance for the roll is symmetric. This is probably because roll angles change vertical cues symmetrically whereas pitch does not. Similar to pitch, the performance for the networks trained on roll (R) and pitch and roll (PR) is maintained over the range of roll angles. As for the pitch, the mean result is near the 75th percentile as the average is skewed by larger outliers in error.

\begin{figure}[h]
    \centering
    \includegraphics[width=0.9\textwidth]{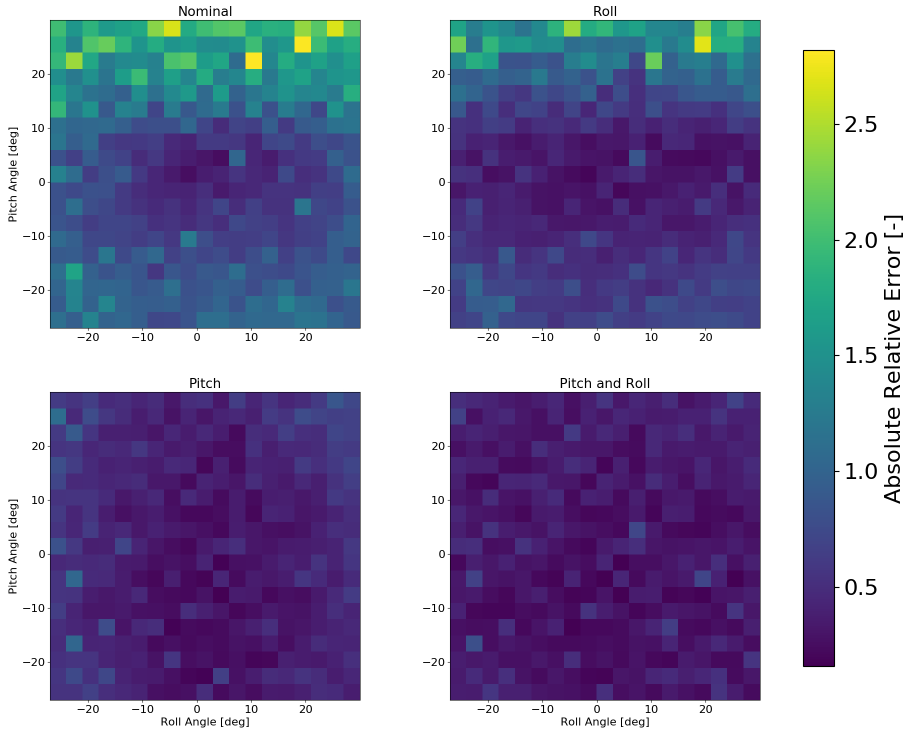}
    \caption{Comparison of performance of networks trained on all four datasets on the pitch and roll (PR) test set as a function of both pitch and roll angle.}
    \label{fig:3d_comparison}
\end{figure}

For a 3D analysis, \autoref{fig:3d_comparison} shows the performance on the pitch and roll (PR) dataset over the range of pitch and roll angle, for 4 networks trained on each of the datasets. The first significant result is that the network trained on the nominal (N) dataset performs far worse, overall, than the other networks. The performance near smaller angles of roll and pitch is good for the network trained on the nominal (N) dataset. However, for larger angles the performance quickly degrades to large errors. This is especially the case for positive pitch angles. The network trained on the roll (R) dataset has a similar shape to that of the nominal, that larger pitch angles degrade performance substatially. However, for each pitch angle, along the range of roll angles, the performance is fairly consistent.

Both the networks trained on the pitch (P) and pitch and roll (PR) seem to have more consistent, and better, performance over the range of rotations. Although, performance on the pitch (P) trained network does seem to slightly worsen when going to extreme roll angles. The performance of the pitch and roll (PR) trained network is constant over the entire range of 2D rotations and is overall the best. From these results, it is apparent that pitch is the more important rotation for improving performance on dynamic platforms. Also, it can be seen that training on both pitch and roll is the most effective for improving performance over the entire range of expected rotations.

